\documentclass{IEEEcsmag}

\usepackage[colorlinks,urlcolor=blue,linkcolor=blue,citecolor=blue]{hyperref}
\expandafter\def\expandafter\UrlBreaks\expandafter{\UrlBreaks\do\/\do\*\do\-\do\~\do\'\do\"\do\-}
\usepackage{upmath,color}

\usepackage{listings}
\usepackage{float}
\usepackage[super]{natbib}
\usepackage{caption}
\usepackage{comment}

\usepackage{orcidlink}
\usepackage{subfigure}
\usepackage{graphicx}
\usepackage{tabularx}
\newcolumntype{Y}{>{\centering\arraybackslash}X}

\jvol{XX}
\jnum{XX}
\paper{8}
\jmonth{April}
\jname{IEEE Intelligent Systems}
\jtitle{IEEE Intelligent Systems}
\pubyear{2024}

\usepackage{hyperref}
\hypersetup{
    colorlinks=true,
    linkcolor=blue,
    filecolor=blue,
    citecolor=green,
    urlcolor=blue,
    }

\begin{document}

\sptitle{Feature Article: Localizing Object Visual Key Field in VQA}

\title{Detect2Interact: Localizing Object Key Field in Visual Question Answering (VQA) with LLMs}

\author{Jialou Wang$^{\orcidlink{0000-0001-8833-4360}}$}
\affil{Northumbria University, NE1 8ST, Newcastle upon Tyne, UK.}

\author{Manli Zhu$^{\orcidlink{0000-0002-8231-5342}}$}
\affil{Northumbria University, NE1 8ST, Newcastle upon Tyne, UK.}

\author{Yulei Li$^{\orcidlink{0000-0003-3579-6179}}$}
\affil{Northumbria University, NE1 8ST, Newcastle upon Tyne, UK.}

\author{{Honglei Li}$^{\orcidlink{0000-0003-3126-0764}}$}
\affil{Northumbria University, NE1 8ST, Newcastle upon Tyne, UK.}

\author{{Longzhi Yang}$^{\orcidlink{0000-0003-2115-4909}}$}
\affil{Northumbria University, NE1 8ST, Newcastle upon Tyne, UK.}

\author{Wai Lok Woo $^{\orcidlink{0000-0002-8698-7605}}$}
\affil{Northumbria University, NE1 8ST, Newcastle upon Tyne, UK.}

\markboth{THEME/FEATURE/DEPARTMENT}{THEME/FEATURE/DEPARTMENT}

\begin{abstract} Localization plays a crucial role in enhancing the practicality and precision of VQA systems. By enabling fine-grained identification and interaction with specific parts of an object, it significantly improves the system's ability to provide contextually relevant and spatially accurate responses, crucial for applications in dynamic environments like robotics and augmented reality. However, traditional systems face challenges in accurately mapping objects within images to generate nuanced and spatially aware responses. In this work, we introduce ``Detect2Interact'', which addresses these challenges by introducing an advanced approach for fine-grained object visual key field detection. First, we use the segment anything model (SAM) to generate detailed spatial maps of objects in images. Next, we use Vision Studio to extract semantic object descriptions. Third, we employ GPT-4's common sense knowledge, bridging the gap between an object's semantics and its spatial map. As a result, Detect2Interact achieves consistent qualitative results on object key field detection across extensive test cases and outperforms the existing VQA system with object detection by providing a more reasonable and finer visual representation.
\end{abstract}

\maketitle

\begin{quote}
\textit{``Only close attention to the fine details of any operation makes the operation first class.'' \\--- J. Willard Marriott}
\end{quote}

\chapteri{T}he task of Visual Question Answering (VQA) aims to jointly reason over visual scenes and natural language inputs, advancing both computer vision and natural language processing domains. This fusion of vision and language enhances human-machine interaction. It allows machines to effectively communicate with humans using natural language instructions, offering intuitive, context-aware visual responses.\cite{Interactive-language,DetGPT2023} By accurately localizing and classifying objects in a scene, object detection allows VQA systems to support intuitive interactions between visual and language, enhancing their utility for a wide range of users. For instance, when posed with a question like ``how can I move the mug?", a VQA system equipped with fine-grained object detection can identify specific components, such as the handle of the mug. This kind of fine-grained detection of specific object parts ensures the response of a VQA system is not only accurate but also actionable for applications ranging from robotics to augmented reality. Human-robot interactions, for example, would benefit greatly from such specificity. A robot that understands the significance of a mug's handle can grip it correctly, preventing spills and potential mishaps. In augmented reality applications,\cite{AR2022} detailed object annotations facilitate accurate tracking of object parts, resulting in a better user experience. 

Existing VQA systems only focus on providing answers in a textual format. \cite{prophet2023} They are good at handling general questions, especially those that require a holistic understanding of an entire image rather than specific objects. As a result, they can only respond with the coarse location of an object like ``top left'' or ``bottom right'' when questions like ``where is the tiger?'' are asked. Recent works,\cite{VisionLLM2023,Kosmos-2} have leveraged vision models to achieve visual grounding by providing the text format of bounding boxes. Nevertheless, such textual responses are neither intuitively interpretable for humans nor operationable for machines. To alleviate this issue, a very recent work proposed MiniGPT-v2,\cite{minigptv2} which explored a task-oriented instruction training scheme, allowing the visual illustration of bounding boxes for instruction-related objects. However, with limited capacity for fine-grained object part detection, this indirect approach primarily identifies objects at a coarse level by relying on the LLaMA-2 language model to generate textual bounding box representations for object positions.\cite{Llama2} In other words, it cannot recognize key fields of objects, which is essential to understanding the affordance of an object, facilitating machine-object interactions.  

Recognizing the limitations of existing systems, we propose Detect2Interact, which bridges the gap between fine-grained object key field detection and VQA. It integrates segmentation and captioning while leveraging the vast knowledge base of large language models to align an object and its location. This enables the system to detect key fields of an object while providing textual descriptions. First, we propose a zero-shot semantic object detection module that segments an input image with SAM\cite{SAM} and extracts descriptions of these segments to understand the structural semantics of the entirety of each object and its components. Second, we present a target object retrieval module that utilizes the common sense knowledge of a large language model (i.e., GPT-4) to map segmented objects with user queries to extract relevant objects and their contextual significance. Third, we introduce a key-field detection module that leverages the GPT-4 to analyze the action such as ``grab'' of an input query identifying the operable part (i.e., key field) of the extracted object. 

In light of the embryonic state of fine-grained object key-field research in VQA, we conduct extensive qualitative experiments to demonstrate the effectiveness and robustness of our proposed system. Experimental results show that Detect2Interact achieves consistent performance in detecting object key fields across different test cases. In addition, by providing reasonable and finer visual representations, Detect2Interact outperforms the existing VQA system MiniGPT-v2,\cite{minigptv2} which is the only work that considers weak fine-grained object detection to the best of our knowledge.

\section{Related Works}

Visual Question Answering (VQA) has been attracting the interest of many researchers since it bridges the gap between visual scenes and language understanding, advancing both computer vision and natural language processing domains. Earlier VQA models\cite{VQA16} often have constrained questions and answers. For example, the method\cite{VQA30} restricts its questions to those with answers from a predefined list of 16 basic colours or 894 object categories. Similarly, the study\cite{VQA16} focuses on questions formed using templates based on a fixed vocabulary of objects and the relationships between them. Later on, the task of free-form and open-ended VQA\cite{VQA2015} was first introduced. Its goal is to accurately answer open-ended questions about a given image using natural language. With the remarkable generalization abilities of LLMs, some studies\cite{MiniGPT-4-2023,prophet2023} have leveraged LLMs for zero-shot VQA by aligning visual information with the common sense knowledge of LLMs. 

The evolution of VQA has been increasingly steering towards more intricate challenges such as contextual reasoning and fine-grained analysis. This advancement is essential as it elevates the system's understanding from merely recognizing objects and their categories to interpreting complex scenes and relationships within an image. This shift is noticeable in how recent studies are pushing the boundaries beyond standard object recognition, venturing into the realms of spatial understanding and interaction among multiple objects within a scene. For example, Desta et al.\cite{ObjectReasoning2018} proposed to combine object detection and reasoning modules to capture high-level, abstract facts extracted from inputs, facilitating reasoning in VQA for better performance. Garg et al. \cite{ObjectQuery2018} introduced a novel method to encode visual information including categorical and spatial information of all objects into object sequences for yes/no VQA tasks, resulting in improved performance compared to baselines. Gupta et al. \cite{ObjectClassification2020} trained a convolutional neural network for object classification that enables it to recognize the features associated with different objects, improving the performance of VQA. Furthermore, Xi et al. \cite{xi2020visual} presented a VQA model based on multi-objective visual relationship detection, incorporating appearance features and relationship predicates for better generalization.

Despite these advancements, current VQA methods struggle to detect or highlight objects based on user queries, and they falter with novel objects, heavily reliant as they are on closed-class object detectors and classifiers. A notable advancement addressing this issue is MiniGPT-v2,\cite{minigptv2} which visualizes bounding boxes around objects in response to specific instructions. This innovative model consolidates reduced visual tokens into the linguistic structure of the large language model LLaMA-2,\cite{Llama2} using a three-layered approach comprising a visual backbone, linear projection layer, and the language model itself. The paper \cite{minigptv2} evaluates the model on diverse vision-language benchmarks and show that it achieves state-of-the-art or comparable performance compared to other models. However, despite MiniGPT-v2's proficiency in translating detailed visual inputs into textual descriptions, it lacks in offering intuitive, interactive elements such as precise object localization and the ability to understand and utilize object affordances. This gap highlights the need for a more advanced approach, one that doesn’t just comprehend but also more effectively and precisely interacts with the visual aspects of a query.

\begin{figure*}[htbp]
    \centering
    \includegraphics[width=\linewidth]{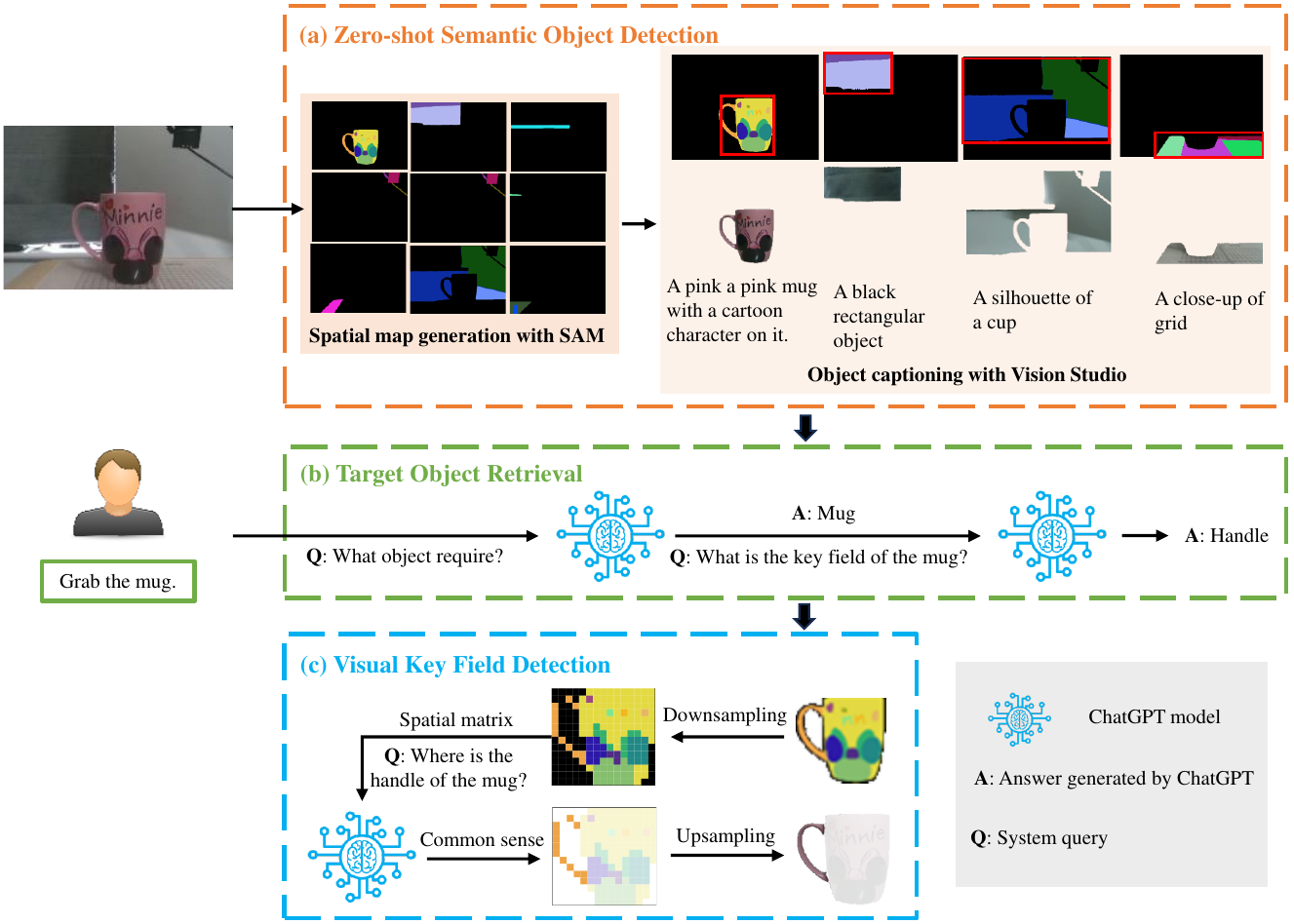}
    \caption{\footnotesize{An overview of our Detect2Interact framework. (a) Given an image, we first adopt SAM \cite{SAM} to segment everything within this image, generating spatial maps of all objects. We then use Vision Studio proposed by Microsoft to obtain objects' semantics. Finally, zero-shot object detection is achieved by combining the spatial maps and object semantics. (b) Given a user query, we utilize the common sense knowledge of ChatGPT to extract the target object (e.g., the ``mug'') and to interpret the user action (e.g., ``grab''). By feeding the extracted information back to ChatGPT, the key-field semantic of the target object (e.g., the ``handle'') is determined. (c) We then feed the spatial matrix of the target object into ChatGPT to recognize its specific visual key field that fits the user action.}}
    \label{fig:flowchart}
\end{figure*}

\section{Overview of Detect2Interact}
In this paper, we propose Detect2Interact, a novel approach focusing on the detection of fine-grained object fields and leveraging object affordances for more enriched interactive VQA applications, thus addressing gaps left by models like MiniGPT-v2. This proposed approach integrates segmentation and captioning while leveraging the knowledge base of large language models to align an object and its location. This enables the system to detect objects in a zero-shot manner and to detect visual fields of an object while providing textual descriptions. As shown in Figure \ref{fig:flowchart}, our framework consists of three components, i.e., a zero-shot semantic object detection module that detects all the objects with semantics, a target object retrieval module that extracts the target object given a user query, and a visual key field detection module that aligns common sense with object spatial matrix recognizing the visual key field that fits the user action. They are detailed in the subsequent sections.

\subsection{Zero-shot Semantic Object Detection}
The Zero-shot Semantic Object Detection (SOD) module aims to detect all the objects with semantics present in a given image in a zero-shot fashion. To achieve this, we take advantage of the state-of-the-art segmentation model SAM \cite{SAM} proposed by Meta and the image analysis model Vision Studio proposed by Microsoft. SAM is an advanced deep-learning model for the task of image segmentation. It is trained on a large amount of segmentation data and leverages both CNN and Transformer networks to extract image features in a hierarchical and multi-scale manner. It can identify and segment any object in an image without training, making it a natural choice for us to identify spatial maps of objects. Vision Studio is a state-of-the-art platform for various computer vision tasks including image captioning. It provides an interface that bridges the gap between complex algorithms and real-world applications, making it a preferred choice for us to obtain the semantics of objects.



\begin{figure}[htbp]
    \centering
    \includegraphics[width=0.6\linewidth]{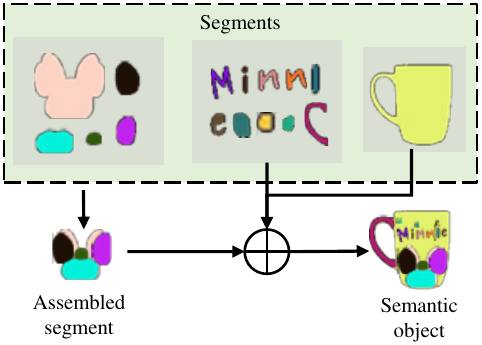}
    \caption{Illustration of object-level composition, in which $\otimes$ represents composition operation.}
    \label{fig:composit}
\end{figure}

\begin{figure*}[htbp]
    \centering
    \includegraphics[width=\linewidth]{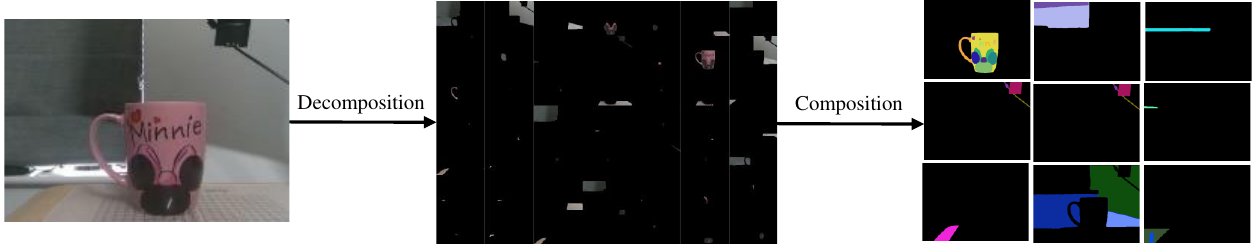}
    \caption{Illustration of object decomposition and composition.}
    \label{fig:segment}
\end{figure*}

Specifically, as depicted in Figure \ref{fig:flowchart} (a), given an input image, we first utilize SAM to generate segmentation masks of all potential objects in the image. Masks with unreasonable sizes (i.e., only include a few pixels) are discarded to reduce noisiness and redundancy. We refer to this process as object decomposition as demonstrated in Figure \ref{fig:segment}. Next, we propose to combine identified segments into semantic objects, similar to assembling different pieces of a jigsaw puzzle as illustrated in Figure \ref{fig:composit}. By examining spatial relationships, similarities, and other contextual cues among the segments, we ensure that these segments represent reasonable and semantically intact objects. We refer to this as object composition in Figure \ref{fig:segment}. Such decomposition and composition design bridges the granular details obtained from SAM with the broader objective of object-level understanding, enabling zero-shot object detection and fine-grained object understanding.

Afterwards, each composited object is associated with a textual descriptor extracted by Vision Studio. This bridges the gap between the spatial object segmentation maps and their semantic language descriptions. Our motivation is that segmentation provides rich spatial information of objects in a finer fashion, and caption offers semantic descriptions, by leveraging the two, we can achieve not only zero-shot object detection but also a fine-grained understanding of objects. Compared to previous zero-shot learning works,\cite{zero-shot2013,2016zero-shot} the main advantage of our system is that it does not need training at all, making it more efficient. Moreover, we overcome the limitation of detecting objects in images using visual representation while providing textual descriptions.


\subsection{Target Object Retrieval}
The Target Object Retrieval (TOR) module aims to extract the target object given a user query, providing further potential interactions with the object. As shown in Figure \ref{fig:flowchart} (b), given a user query such as ``Grab the mug", our system utilizes the common sense knowledge of GPT-4 to look through a list of object descriptors extracted from the SOD module. The descriptors might include entries such as ``A pink mug with a cartoon character on it." and ``A black rectangular object.". By mapping the user's query with these descriptors, GPT-4 determines the target object or a set of target objects. For instance, in response to the query ``Grab the mug", GPT-4 returns ``A pink mug with a cartoon character on it." along with its according id in the object list. More importantly, our approach goes beyond the capability of identifying relevant objects. It seeks to understand the fine-grained semantics of the target object that the user is requesting. When posed with subsequent questions like ``Which part of the mug may solve the request?", GPT-4 recognizes that the ``Handle" of the mug is the key component. Leveraging the capabilities of GPT-4 in such design allows the alignment between the segmented visual objects and the user's intentions. As a result, our system can not only identify objects but also understand their fine-grained significance in context, setting the foundation for sophisticated interactions in various applications.

\subsection{Visual Key Field Detection}
The Visual Key Field Detection (VKFD) module aims to identify the key part of the extracted target object that aligns with the user query, facilitating object interactions. As demonstrated in Figure \ref{fig:flowchart} (b), the original spatial segmentation map of the target object is meticulously downscaled to a shape with a longer side equal to 20 pixels to reduce its size. Since GPT-4 does not provide an interface for image input, we transform the spatial map to a spatial matrix, wherein each constituent element delineates the segment number corresponding to its respective pixel from the downscaled map. We then feed this numerical matrix into the GPT-4 model. By harnessing the model's expansive semantic knowledge base, it identifies the segment correlating with the key field in the query. For example, the model identifies the handle segment in the spatial matrix as synonymous with the ``handle of the mug". By upsampling the spatial map representation, the key field ``handle'' is highlighted in the original object ``mug''. This facilitates the research of more intuitive interactions between visual scenes and textual language by highlighting a specific object or its visual key fields while providing textual answers, enhancing the utility of VQA systems for more general AI applications.

\subsection{Prompt Details}
In this section, we offer a detailed breakdown of the prompt for processing the query 'Draw on the image where can I kick the door open?'. Along with the query, we also input a list containing the semantic information of objects and their spatial maps. In the given example, this information is denoted as `a black door with a handle, [2, 167, 1, 400]'. In the initial stage, we provide the query to GPT-4, probing whether the provided object information suffices for answering the question. GPT-4, upon evaluation, determines that the existing information does not adequately address the query. However, it recognizes that the `door' is the pertinent object for subsequent processing. Now, we introduce the spatial segment matrix that assists GPT-4 in highlighting the exact segments that would accurately respond to the posed question.

Q:Draw on the image where can I kick the door open?
\begin{figure}[h]
    \centering
    \includegraphics[width=0.2\textwidth]{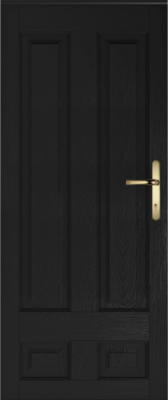}
    \includegraphics[width=0.2\textwidth]{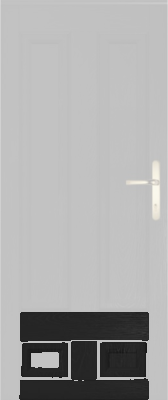}
    \caption{Visual input (left) and output (right).}
    \label{fig:tw image}
\end{figure}

\begin{figure}[ht!]
    \includegraphics[width=0.03\textwidth]{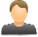}
    \label{fig:user.jpg}
\end{figure}

\begin{lstlisting}[breaklines=true]
[{
`role': `system', 
`content': `You are an expert in logic, a user asked a question regarding an image about [a close-up of a black door]there is a list of objects in the image with it's name and position.
Object list
[id] 0[Description]:a black door with a handle, Position:[2, 167, 1, 400]

Please answer follow questions 
1. Do you have enough information to answer the question? 
2. What object do you think leads us the reveal the answer? 
3. What question are you going to ask the object which will get the answer for the question?

put the answer in the JSON format as following, and must follow the format

``{``Answer:``Yes"/``No",
``Reply":``Oral reply to the question",
``Objects name":[[object id,question to this object],[object id,question to this object] ...,],          
``Position:[[x1,y1,x2,y2],[x1,y1,x2,y2]]" '},
{`role': `user', `content': `Question: where can I kick the door open?'}]
\end{lstlisting}

\begin{figure}[ht!]
    \includegraphics[width=0.05\textwidth]{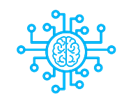}
    \label{fig:ai}
\end{figure}

\begin{lstlisting}[breaklines=true]
{
``Answer":``No", 
``Reply":``The image only provides information about a black door with a handle, but it does not specify any region where you can kick open the door.",
``Objects name":[[0,``Can you specify the region where you can be kicked open?"]],
``Position":[[2, 167, 1, 400]]
}
\end{lstlisting}

\begin{figure}[ht!]
    \includegraphics[width=0.03\textwidth]{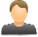}
    \label{fig:user.jpg2}
\end{figure}

\begin{lstlisting}[breaklines=true]
[{
`role': `system', 
`content': `You are a [a black door with a handle]. you are provided a pixel matrix representation of the object, the value for each pixel is the segment id of this pixel, 0 means this pixel is not the part of the object please answer the following question:
1. What do you want to say to the user?
2. Do you think one to more whole single segment fulfill the question? 
3. if so, what is the segment(s) number    
4. if not, what is the area fulfill the question?           

put the answer in the following Json format
{``answer":``Oral answer to the question",
``whole segments":``Yes or No",
``which segment": [segments number],
``target position":[[x1,y1],[x1,y1,x2,y2] ...] ,the relevant position presented by 1 pair of xy for a point(x is width,y is height), 2 pairs for rectangle(x1,y1 is the top left,x2,y2 is the bottom right).r}'}, 
{`role': `user', 
`content': 
`Question:Can you show me the specific region where I can kick open the door?'}, 
{`role': `user', 
`content': `Shape matrix:
[
[1 1 1 1 1 1 1 1 1 0]
[1 1 1 1 1 1 1 1 1 1]
[1 1 1 1 1 1 1 1 1 1]
[1 1 1 1 1 1 1 1 1 1]
[1 1 1 1 1 1 1 1 1 1]
[1 1 1 1 1 1 1 1 1 1]
[1 1 1 1 1 1 1 1 1 1]
[1 1 1 1 1 1 1 1 1 1]
[1 1 1 1 1 1 1 1 1 1]
[1 1 1 1 1 1 1 1 1 1]
[1 1 1 1 1 1 1 8 0 1]
[1 1 1 1 1 1 1 1 0 1]
[1 1 1 1 1 1 1 1 1 1]
[1 1 1 1 1 1 1 1 1 1]
[1 1 1 1 1 1 1 1 1 1]
[1 1 1 1 1 1 1 1 1 1]
[1 1 1 1 1 1 1 1 1 1]
[1 2 2 2 2 2 2 2 1 1]
[1 2 2 2 2 2 2 2 1 1]
[1 4 4 4 3 5 5 5 1 1]
[1 4 4 4 3 5 5 5 1 1]
[1 7 7 7 7 7 7 7 1 1]]
Text and position:[]'}]
\end{lstlisting}

\begin{figure}[ht!]
    \includegraphics[width=0.05\textwidth]{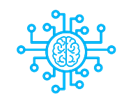}
    \label{fig:ai2}
\end{figure}

\begin{verbatim}
{
`answer': `The region where the door can
be kicked open is at the bottom half of 
the door.', 
`whole segments': `Yes',
`which segment': [2, 3, 4, 5, 7], 
`target position': []}
\end{verbatim}

\section{Qualitative Evaluation}
In light of the embryonic state of fine-grained object key-field research in VQA, we conduct qualitative experiments to evaluate the effectiveness and robustness of the proposed Detect2Interact. We mainly compare it with the very recent MiniGPT-v2 \cite{minigptv2} that considers fine-grained object detection. In the following sections, we showcase multiple test cases across different tasks. These cases not only demonstrate the versatility and robustness of our system in comparison to MiniGPT-v2 but also highlight its adeptness at interpreting and accurately responding to user queries.


\begin{figure}[htbp]
    \centering
    \includegraphics[width=\linewidth]{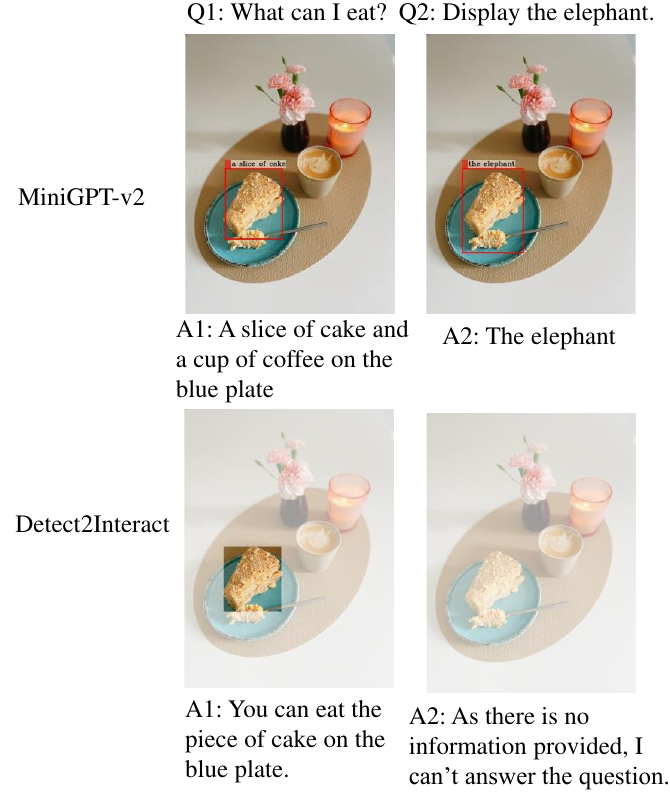}
    \caption{Comparison with MiniGPT-v2 \cite{minigptv2} on zero-shot object detection.}
    \label{fig:od}
\end{figure}

\begin{figure*}[htbp]
    \centering
    \includegraphics[width=\linewidth]{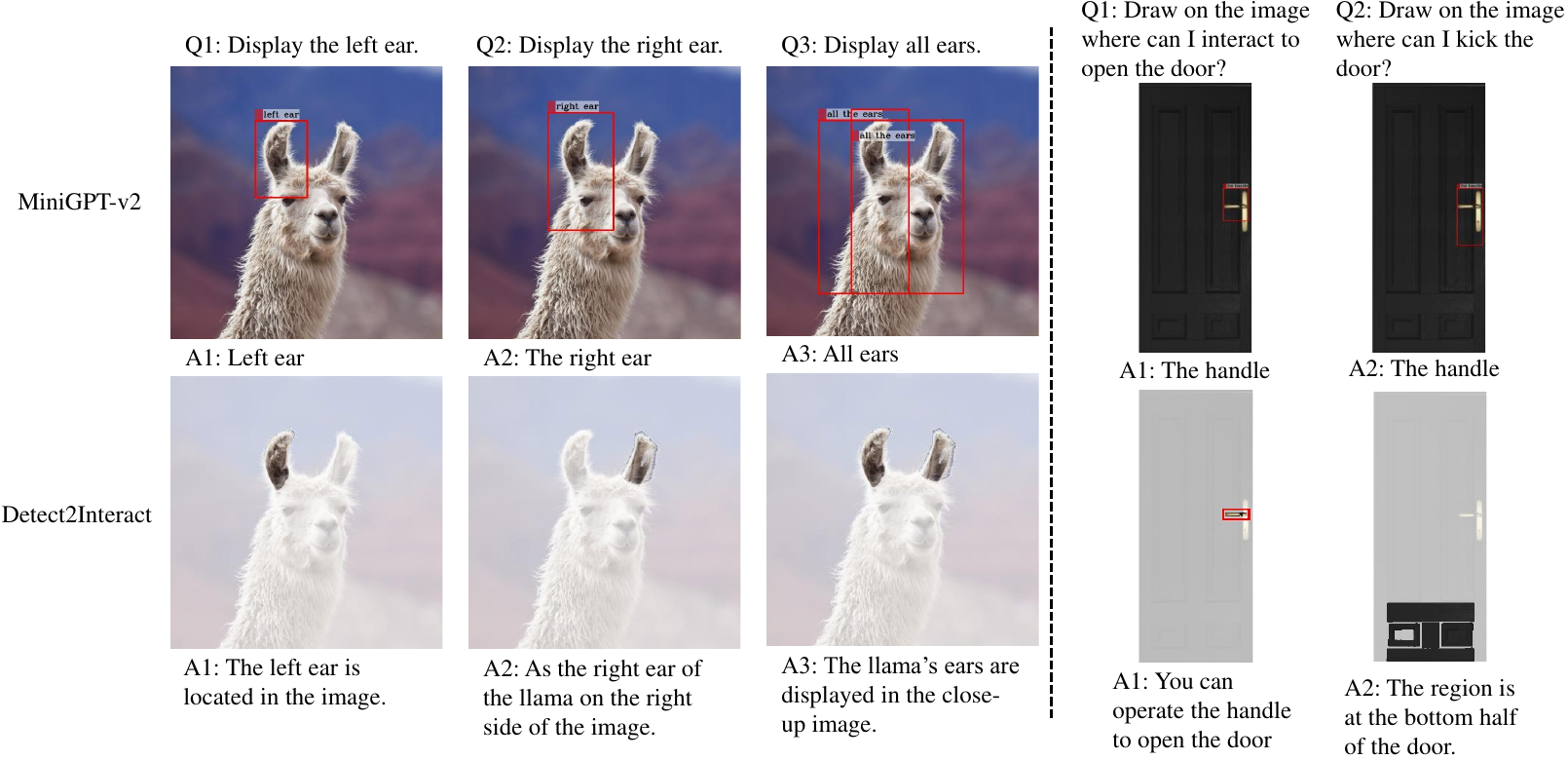}
    \caption{Comparison with MiniGPT-v2 \cite{minigptv2} in VQA. Various questions are tested on both the same and different images. Note that we added a red bounding box to the output when it's too small for better readability.}
    \label{fig:cases}
\end{figure*}

\begin{figure*}[htbp]
    \centering
    \includegraphics[width=\linewidth]{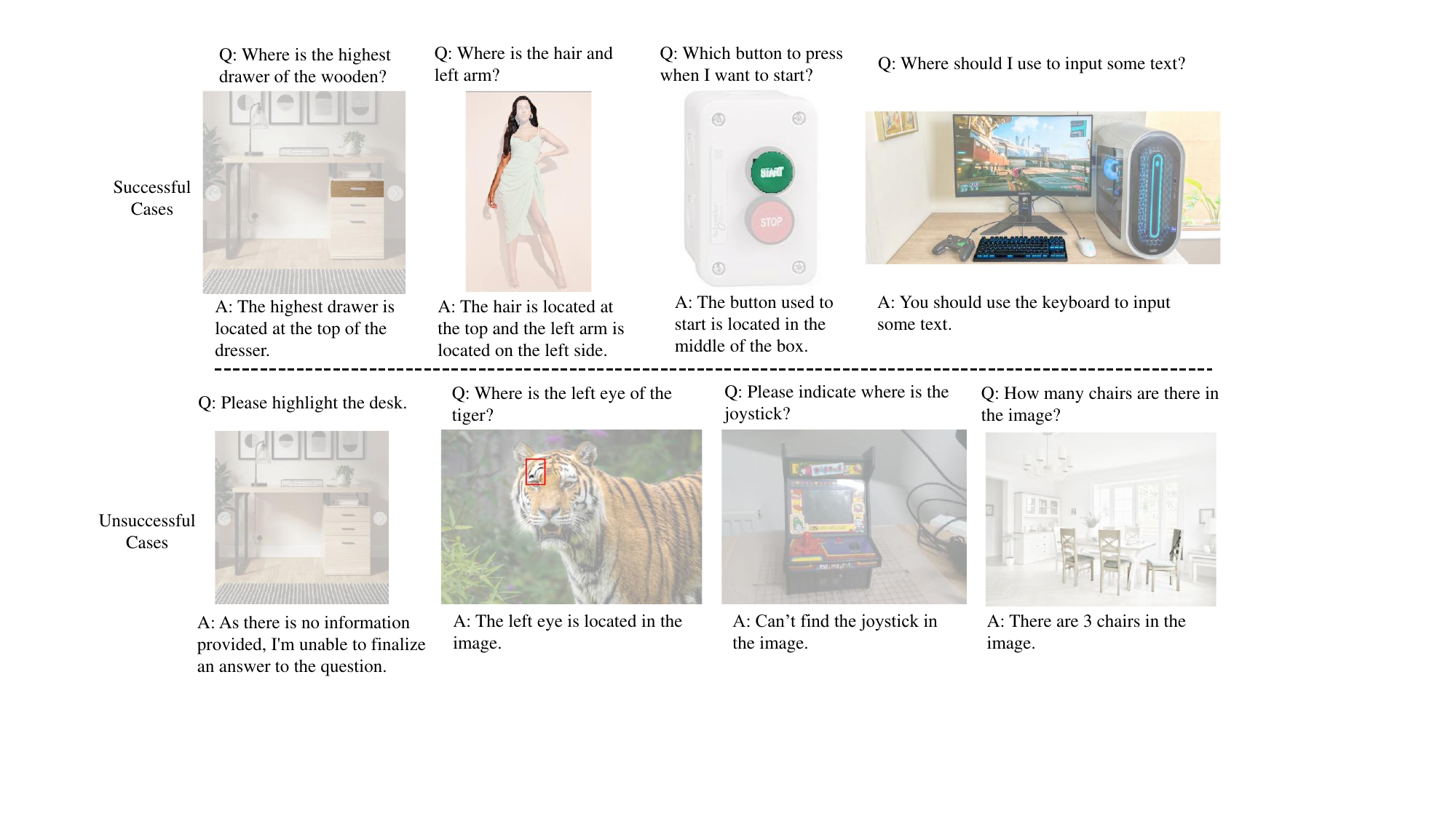}
    \caption{Qualitative results. The upper row showcases the effectiveness of Detect2Interact, while the lower row depicts failure cases.}
    \label{fig:failure}
\end{figure*}

\subsection{System Settings}
Our proposed Detect2Interact is developed using Python with the Tkinter library, providing an intuitive interface depicted in the demonstrations. Users can either upload an image or capture one in real-time, and post their query. The system then offers both textual feedback and visual highlights of the relevant visual parts in the image to address the user's query comprehensively.

We benchmarked our system against MiniGPT-v2\cite{minigptv2}, which offers a comparable service via a web interface accessible at \href{https://minigpt-v2.github.io}{https://minigpt-v2.github.io}. Similar to our platform, MiniGPT-v2 provides users with interactive features for image queries and responses.

\subsection{Test Cases}

\begin{enumerate}
\item \textbf{Object Part Detection and Visual Key Field Detection} In this work, our primary focus is to detect the visual key field for an object in VQA and fine-grained object part detection. As there is limited existing work researching in this area, we compare Detect2Interact with the closest work MiniGPT-v2 \cite{minigptv2} in Figure \ref{fig:cases}. For the fine-grained object part detection, our system's precision stood out distinctly. When asked to highlight specific parts of an object, such as a llama's left or right ear, our system demonstrated consistency and precision. It was able to correctly identify and represent the concrete mask of interest. In contrast, MiniGPT-v2's bounding box representation seemed to be too generic, especially when asked to ``Display all ears," offering a broader and vague box rather than delineating the exact regions. Furthermore, when specifically queried about the right ear, MiniGPT-v2 erroneously highlighted an incorrect area, misrepresenting the accurate location of the ear. More importantly, for the object visual key field detection, MiniGPT-v2's limitations became overtly manifest. It fails to differentiate between the commands ``open the door" and ``kick open the door", producing identical outputs fixated on the door handle.  On the other hand, our system demonstrated nuanced understanding, astutely mapping user queries to the precise operational points on the object, thus validating its superior adaptability and contextual awareness.

\item \textbf{Zero-shot Object Detection} Beyond visual key field detection, we also found that Detect2Interact has a potential application in zero-shot object detection. We show some qualitative results comparing with MiniGPT-v2 in Figure \ref{fig:od}. We can see that both methods can accurately detect the ``cake'' in the image. When the query involves a non-existing object in the image, MiniGPT-v2 fails to give a reasonable answer and tends to fabricate a visual tag for it, while our Detect2Interact can provide the correct answer, i.e., not able to answer the question.

\item \textbf{More Qualitative Results} The examples illustrated in Figure \ref{fig:failure} showcase the framework's capability to accurately locate the fine-grained object within various images. However, the unsuccessful examples in the lower row in Figure \ref{fig:failure} highlight four primary limitations. First, Detect2Interact fails to detect an object with subtle appearance visibility as depicted in the first column. Second, it struggles to identify segments when the target is too small as shown in the second column. Third, it fails to detect less common objects as demonstrated in the third column. Fourth, it detects only more visible objects when there is severe occlusion, as showcased in the last column.
\end{enumerate}

\subsection{Limitation}
Our proposed Detect2Interact localizes interactive and key fields of an object by using the common sense of ChatGPT and pixel-level segmentation of SAM. Despite Detect2Interact demonstrating effective capabilities, it presents a few notable limitations. First, our predominant emphasis on object segmentation might have unintentionally minimized the importance of colour features, potentially causing the system to overlook the crucial distinction that colour can bring in object identification. Second, we downscale input images to a substantially diminished size (20 pixels on the long side) due to the constraints of the GPT-4 API, which could result in the loss of finer details and make it challenging to recognize and respond to smaller objects. Third, Detect2Interact struggles with less common parts (e.g., the joystick of a game machine) of objects due to the limitation of spatial reasoning in GPT-4 itself. GPT-4 is mainly trained on a wide variety of text and images, and we believe such data might not provide enough context or examples for the model to learn complex spatial relationships accurately (\href{https://platform.openai.com/docs/guides/vision/limitations}{GPT-4 limitations}.) Fourth, our system might fail in scenes with severe occlusion due to the reliance on segmentation. Occluded objects, such as a chair positioned behind a table, can be challenging for the system to accurately caption or distinguish its specific segments. Last, since our methodology employs the GPT-4 API, the response time of the API significantly influences the overall execution time. Our experimental observations indicate that the speed of the GPT-4 API can vary at different times and days. Consequently, the execution times recorded in our study may not fully reflect the true performance capabilities of our system in different operational contexts.

\section{Conclusion}
In this paper, we have proposed Detect2Interact, a novel VQA system with fine-grained object parts detection. By leveraging the advantage of SAM for fine-grained object segmentation and Vision Studio for segment caption, Detect2Interact can perform various zero-shot vision-language tasks, including object detection, fine-grained object detection, and key field detection. Extensive test cases have demonstrated the effectiveness and consistency of our system. It facilitates the research of more intuitive interactions between visual scenes and textual language by highlighting a specific object or its key fields while providing textual answers, resulting in a better multi-modal fusion. Importantly, the proposed object key-field detection provides a deeper understanding of the affordance of an object in addition to the global scene. It has a high impact on human-machine applications, such as human-robot interaction and augmented reality. Given the limited benchmark datasets on the new task of object parts detection, we are planning to develop such a dedicated dataset in our future work, advancing VQA systems and their applications. \vspace*{-8pt}


\bibliographystyle{unsrtnat}  
\bibliography{references}

\begin{thebibliography}{19}
\providecommand{\natexlab}[1]{#1}
\providecommand{\url}[1]{\texttt{#1}}
\expandafter\ifx\csname urlstyle\endcsname\relax
  \providecommand{\doi}[1]{doi: #1}\else
  \providecommand{\doi}{doi: \begingroup \urlstyle{rm}\Url}\fi

\bibitem[Lynch et~al.(2023)Lynch, Wahid, Tompson, Ding, Betker, Baruch,
  Armstrong, and Florence]{Interactive-language}
Corey Lynch, Ayzaan Wahid, Jonathan Tompson, Tianli Ding, James Betker, Robert
  Baruch, Travis Armstrong, and Pete Florence.
\newblock Interactive language: Talking to robots in real time.
\newblock \emph{IEEE Robotics and Automation Letters}, pages 1--8, 2023.
\newblock \doi{10.1109/LRA.2023.3295255}.

\bibitem[Pi et~al.(2023)Pi, Gao, Diao, Pan, Dong, Zhang, Yao, Han, Xu, and
  Zhang]{DetGPT2023}
Renjie Pi, Jiahui Gao, Shizhe Diao, Rui Pan, Hanze Dong, Jipeng Zhang, Lewei
  Yao, Jianhua Han, Hang Xu, and Lingpeng Kong~Tong Zhang.
\newblock Detgpt: Detect what you need via reasoning.
\newblock \emph{arXiv preprint arXiv:2305.14167}, 2023.

\bibitem[Apostolopoulos et~al.(2022)Apostolopoulos, Andronas, Fourtakas, and
  Makris]{AR2022}
George Apostolopoulos, Dionisis Andronas, Nikos Fourtakas, and Sotiris Makris.
\newblock Operator training framework for hybrid environments: an augmented
  reality module using machine learning object recognition.
\newblock \emph{Procedia CIRP}, 106:\penalty0 102--107, 2022.

\bibitem[Shao et~al.(2023)Shao, Yu, Wang, and Yu]{prophet2023}
Zhenwei Shao, Zhou Yu, Meng Wang, and Jun Yu.
\newblock Prompting large language models with answer heuristics for
  knowledge-based visual question answering.
\newblock In \emph{Proceedings of the IEEE/CVF Conference on Computer Vision
  and Pattern Recognition}, pages 14974--14983, 2023.

\bibitem[Wang et~al.(2023)Wang, Chen, Chen, Wu, Zhu, Zeng, Luo, Lu, Zhou, Qiao,
  et~al.]{VisionLLM2023}
Wenhai Wang, Zhe Chen, Xiaokang Chen, Jiannan Wu, Xizhou Zhu, Gang Zeng, Ping
  Luo, Tong Lu, Jie Zhou, Yu~Qiao, et~al.
\newblock Visionllm: Large language model is also an open-ended decoder for
  vision-centric tasks.
\newblock \emph{arXiv preprint arXiv:2305.11175}, 2023.

\bibitem[Peng et~al.(2023)Peng, Wang, Dong, Hao, Huang, Ma, and Wei]{Kosmos-2}
Zhiliang Peng, Wenhui Wang, Li~Dong, Yaru Hao, Shaohan Huang, Shuming Ma, and
  Furu Wei.
\newblock Kosmos-2: Grounding multimodal large language models to the world.
\newblock \emph{arXiv preprint arXiv:2306.14824}, 2023.

\bibitem[Chen et~al.(2023)Chen, Zhu, Shen, Li, Liu, Zhang, Krishnamoorthi,
  Chandra, Xiong, and Elhoseiny]{minigptv2}
Jun Chen, Deyao Zhu, Xiaoqian Shen, Xiang Li, Zechun Liu, Pengchuan Zhang,
  Raghuraman Krishnamoorthi, Vikas Chandra, Yunyang Xiong, and Mohamed
  Elhoseiny.
\newblock Minigpt-v2: large language model as a unified interface for
  vision-language multi-task learning, 2023.

\bibitem[Touvron et~al.(2023)Touvron, Martin, Stone, Albert, Almahairi, Babaei,
  Bashlykov, Batra, Bhargava, Bhosale, et~al.]{Llama2}
Hugo Touvron, Louis Martin, Kevin Stone, Peter Albert, Amjad Almahairi, Yasmine
  Babaei, Nikolay Bashlykov, Soumya Batra, Prajjwal Bhargava, Shruti Bhosale,
  et~al.
\newblock Llama 2: Open foundation and fine-tuned chat models.
\newblock \emph{arXiv preprint arXiv:2307.09288}, 2023.

\bibitem[Kirillov et~al.(2023)Kirillov, Mintun, Ravi, Mao, Rolland, Gustafson,
  Xiao, Whitehead, Berg, Lo, et~al.]{SAM}
Alexander Kirillov, Eric Mintun, Nikhila Ravi, Hanzi Mao, Chloe Rolland, Laura
  Gustafson, Tete Xiao, Spencer Whitehead, Alexander~C Berg, Wan-Yen Lo, et~al.
\newblock Segment anything.
\newblock \emph{arXiv preprint arXiv:2304.02643}, 2023.

\bibitem[Geman et~al.(2015)Geman, Geman, Hallonquist, and Younes]{VQA16}
Donald Geman, Stuart Geman, Neil Hallonquist, and Laurent Younes.
\newblock Visual turing test for computer vision systems.
\newblock \emph{Proceedings of the National Academy of Sciences}, 112\penalty0
  (12):\penalty0 3618--3623, 2015.

\bibitem[Malinowski and Fritz(2014)]{VQA30}
Mateusz Malinowski and Mario Fritz.
\newblock A multi-world approach to question answering about real-world scenes
  based on uncertain input.
\newblock \emph{Advances in neural information processing systems}, 27, 2014.

\bibitem[Antol et~al.(2015)Antol, Agrawal, Lu, Mitchell, Batra, Zitnick, and
  Parikh]{VQA2015}
Stanislaw Antol, Aishwarya Agrawal, Jiasen Lu, Margaret Mitchell, Dhruv Batra,
  C~Lawrence Zitnick, and Devi Parikh.
\newblock Vqa: Visual question answering.
\newblock In \emph{Proceedings of the IEEE international conference on computer
  vision}, pages 2425--2433, 2015.

\bibitem[Zhu et~al.(2023)Zhu, Chen, Shen, Li, and Elhoseiny]{MiniGPT-4-2023}
Deyao Zhu, Jun Chen, Xiaoqian Shen, Xiang Li, and Mohamed Elhoseiny.
\newblock Minigpt-4: Enhancing vision-language understanding with advanced
  large language models.
\newblock \emph{arXiv preprint arXiv:2304.10592}, 2023.

\bibitem[Desta et~al.(2018)Desta, Chen, and Kornuta]{ObjectReasoning2018}
Mikyas~T Desta, Larry Chen, and Tomasz Kornuta.
\newblock Object-based reasoning in vqa.
\newblock In \emph{2018 IEEE Winter Conference on Applications of Computer
  Vision (WACV)}, pages 1814--1823. IEEE, 2018.

\bibitem[Garg and Srivastava(2018)]{ObjectQuery2018}
Shivam Garg and Rajeev Srivastava.
\newblock Object sequences: encoding categorical and spatial information for a
  yes/no visual question answering task.
\newblock \emph{IET Computer Vision}, 12\penalty0 (8):\penalty0 1141--1150,
  2018.

\bibitem[Gupta et~al.(2020)Gupta, Garg, Deshmane, Singh, and
  Agarwal]{ObjectClassification2020}
Akshra Gupta, Manas Garg, Shrikant Deshmane, Parikshit~Kishor Singh, and Basant
  Agarwal.
\newblock Object-based classification for visual question answering.
\newblock In \emph{2020 3rd International Conference on Emerging Technologies
  in Computer Engineering: Machine Learning and Internet of Things (ICETCE)},
  pages 98--104. IEEE, 2020.

\bibitem[Xi et~al.(2020)Xi, Zhang, Ding, and Wan]{xi2020visual}
Yuling Xi, Yanning Zhang, Songtao Ding, and Shaohua Wan.
\newblock Visual question answering model based on visual relationship
  detection.
\newblock \emph{Signal Processing: Image Communication}, 80:\penalty0 115648,
  2020.

\bibitem[Lampert et~al.(2013)Lampert, Nickisch, and Harmeling]{zero-shot2013}
Christoph~H Lampert, Hannes Nickisch, and Stefan Harmeling.
\newblock Attribute-based classification for zero-shot visual object
  categorization.
\newblock \emph{IEEE transactions on pattern analysis and machine
  intelligence}, 36\penalty0 (3):\penalty0 453--465, 2013.

\bibitem[Xian et~al.(2016)Xian, Akata, Sharma, Nguyen, Hein, and
  Schiele]{2016zero-shot}
Yongqin Xian, Zeynep Akata, Gaurav Sharma, Quynh Nguyen, Matthias Hein, and
  Bernt Schiele.
\newblock Latent embeddings for zero-shot classification.
\newblock In \emph{Proceedings of the IEEE conference on computer vision and
  pattern recognition}, pages 69--77, 2016.

\end{thebibliography}

\begin{IEEEbiography}{Jialou Wang}{\,}is a doctoral candidate at Department of Computer and Information Sciences, Northumbria University, Newcastle upon Tyne, UK.His current research interests include Large Language Model, Computer Vision and Robotic. Wang received his Bachlor degree in computer science from Victoria University of Wellington,New Zealand. Contact him at jialou.wang@northumbria.ac.uk.\vspace*{8pt}
\end{IEEEbiography}

\begin{IEEEbiography}{Manli Zhu}{\,} is a PhD candidate with the Department of Computer and Information Sciences at Northumbria University, Newcastle upon Tyne, UK. Her current research interests include computer vision, interaction modelling, and human motion analysis. She received her master's degree in the School of Computer Science and Information Security from Guilin University of Electronic Technology, China. Contact her at manli.zhu@northumbria.ac.uk.\vspace*{8pt}
\end{IEEEbiography}

\begin{IEEEbiography}{Dr. Yulei Li} {\,} currently a Lecturer in Computer and Information Science at Northumbria University, UK, specializes in the impactful application of Large Language Models (LLMs), AI, and machine learning in diverse business environments. His work primarily concentrates on leveraging these advanced technologies to drive significant improvements and innovations in industry practices, demonstrating the profound effects of AI on operational efficiency and strategic development. Contact him at yulei3.li@northumbria.ac.uk.
\end{IEEEbiography}

\begin{IEEEbiography}{Honglei Li}{\,} is a senior lecturer in Enterprise Information Systems in the Department of Computer and Information Sciences at Northumbria University, UK.expertise in end-user information technology adoption, particularly for virtual communities, smart cities, online recommendation systems, and artificial intelligence.She received her PhD degree from Chinese University of Hong Kong,HK,CN. Contact her at honglei.li@northumbria.ac.uk
\end{IEEEbiography}

\begin{IEEEbiography}{Longzhi Yang}{\,} is a Professor of AI and Computer Science, and the Head of Education with the Department of Computer and Information Sciences at Northumbria University in Newcastle upon Tyne, UK. His research interests include artificial intelligence, intelligent systems, and robotics. He is a Fellow of British Computer Society, a Senior Fellow of Higher Education Academy, and a Senior Member of IEEE. He received his PhD degree from the University of Wale, Aberystwyth, UK. Contact him at longzhi.yang@northumbria.ac.uk.
\end{IEEEbiography}

\begin{IEEEbiography}{Wai Lok Woo}{\,} is a Professor of AI and Computer Science, the Faculty Postgraduate Research Director (Engineering and Environment), and Head of Research Cluster for Artificial Intelligence and Digital Technology. His research interests include artificial intelligence, machine learning, and data mining. He is a Fellow of the Institution Engineering Technology (IET) and Institution of Electrical and Electronic Engineering (IEEE). He is the corresponding author of this article. Contact him at wailok.woo@northumbria.ac.uk.
\end{IEEEbiography}

\end{document}